\documentclass{bmvc2k}

\usepackage{float}
\usepackage{xcolor}
\usepackage{multirow}

\title{Scene Text Recognition with Semantics}

\addauthor{Joshua Cesare Placidi*}{jplacidi@ed.ac.uk}{1}
\addauthor{Yishu Miao}{y.miao20@imperial.ac.uk}{2}
\addauthor{Zixu Wang}{zixu.wang@imperial.ac.uk}{2}
\addauthor{Lucia Specia}{l.specia@imperial.ac.uk}{2}

\addinstitution{
 School of Informatics\\
 University of Edinburgh\\
 Edinburgh, UK
}
\addinstitution{
 Department of Computing\\
 Imperial College London\\
 London, UK
}

\runninghead{Placidi, Miao, Wang, Specia}{Scene Text Recognition with Semantics}

\usepackage{booktabs}

\begin{document}

\maketitle

\begin{abstract}
    Scene Text Recognition (STR) models have achieved high performance in recent years on benchmark datasets where text images are presented with minimal noise.
    Traditional STR recognition pipelines take a cropped image as sole input and attempt to identify the characters present. 
    This infrastructure can fail in instances where the input image is noisy or the text is partially obscured.
    This paper proposes using semantic information from the greater scene to contextualise predictions.
    We generate semantic vectors using object tags and fuse this information into a transformer-based architecture. 
    The results demonstrate that our multimodal approach yields higher performance than traditional benchmark models, particularly on noisy instances.
\end{abstract}

\section{Introduction}
\label{sec:intro}

Optical Character Recognition (OCR) \cite{Islam_ocr_survey, Sabu_OCR_survey} is the task of converting an image of text into machine encoded characters and is widely used in industrial application such as scanning documents and image searching.
Scene Text Recognition (STR) \cite{Baek_2019_ICCV, Ye_STR_survey, Liu_STR_survey, Long_STR_survey}, a subset of the OCR literature, is concerned with detecting text in unpredictable and imperfect environments, so called ``natural images''.
Applications include street sign recognition, assisting the blind or partially sighted, and automatic text translation among others.
STR can be broken down into two stages, detecting text instances in an image scene and recognising the text within localised instances, this work focuses on the latter.
The primary challenge of STR is the inherent variety and diversity in input images.
Text can take many different forms with an enormity of font variations, character sizes, and type-spacing.
Furthermore, models must be invariant to different backgrounds, lighting conditions, resolutions, and distortions.
The combination of these factors yield a difficult challenge for designing robust STR systems.


\begin{figure}[H]
\centering
\begin{tabular}{ccc}
\includegraphics[width=0.3\textwidth]{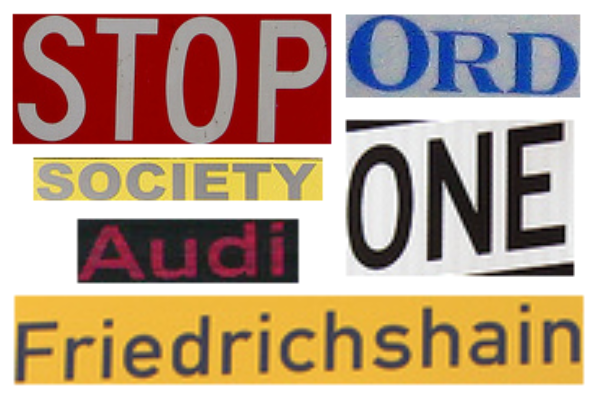} &
\includegraphics[width=0.3\textwidth]{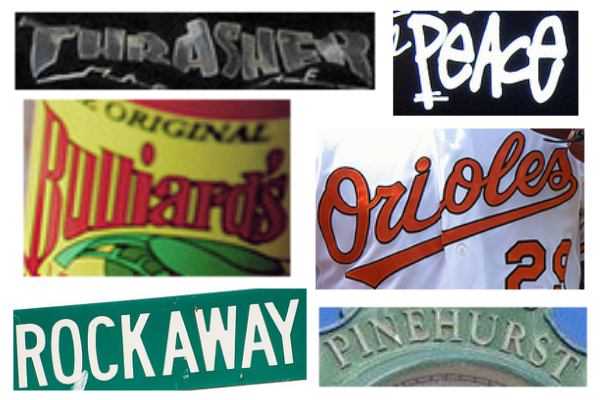} &
\includegraphics[width=0.3\textwidth]{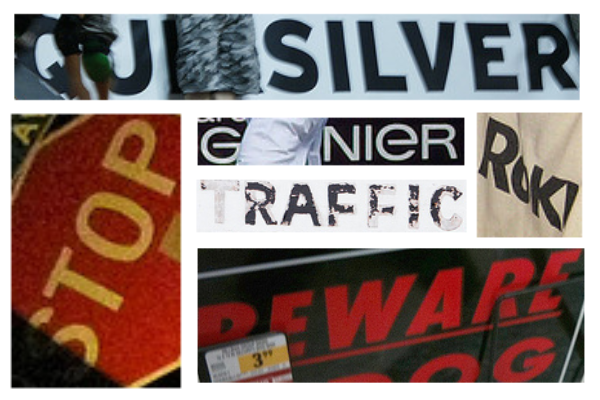} \\
(a) Regular Text & (b) Irregular Text & (c) Hard Text
\end{tabular}
\caption{Comparison of STR text types}
\label{fig:STR text type comparison}
\end{figure}

Generally STR tasks and datasets can be divided into two categories: regular and irregular text.
Regular text contains little noise, with text being level and of constant size.
Irregular text tasks discard these assumptions, containing large variations in font, size, curvature and other distortions.
In this work we are interested in more challenging scenarios where text contains significant distortion and obstruction to the point that some characters can be completely obscured from the image, we refer to this as hard text.
Figure \ref{fig:STR text type comparison} shows examples of each text category.


Previous STR works are focused on regular text and irregular text \cite{Mishra_BMVC, shi2016, Baek_2019_ICCV, scatter, Lu2021MASTER, STN-OCR, OCR-on-the-go}. 
While high performance has been achieved on these tasks, past approaches struggle with imperfect and noisy input images (hard text). In this work we look at common failure cases in STR models and how models can be designed to be more robust to them.
Conventional approaches to STR \cite{Baek_2019_ICCV, STAR-Net, Wang2017_GRCNN, Rosetta} only consider the cropped text image when making predictions and discard the larger but useful scene, which implicitly provides information that can aid STR, particularly on hard text samples.

We propose using visual semantic information from image scenes to contextualised and aid predictions.
Providing the model with access to scene information can instil probability priors which can be utilised in cases of high uncertainty, such as differentiating between the character `S' and digit `5'.
We build upon the design and training framework proposed \cite{Baek_2019_ICCV}, but replace their proposed RNN structure with a transformer architecture to encode and decode visual feature sequences.
The model takes semantic information and cropped images as input. Initially, these are treated as separate modalities and processed as such. We then experiment with different fusion strategies to combine the input streams and generate final predictions.
The experimental results show that our multimodal architecture achieves higher performance than standard STR pipelines, particularly on noisy and partially obstructed images.

Our contributions can be summarised as follows: 
(1) A new multimodal approach to STR tasks using semantic information from the greater image scene.
(2) A transformer-based encoder-decoder architecture which significantly outperforms previous RNN structures.
(3) State-of-the-art results on two full scene datasets: COCOText \cite{veit2016cocotext} and the recently released TextOCR \cite{singh2021textocr}.
(4) A multimodal SemanticSTR model which is more robust to hard text instances.

\section{Related Work}


\paragraph{Scene Text Recognition} 
Modern STR approaches are based on neural models, as detailed in various surveys  \cite {Baek_2019_ICCV, Ye_STR_survey, Long_STR_survey, Liu_STR_survey}.
Early work mainly used bottom-up approaches, classifying characters individually using a sliding window \cite{Wang_STR_2011, Mishra_BMVC,  phan_STR_2013}.
Characters were then combined into a complete word using dynamic programming or graph models.
This method required character-level annotations, which are time and cost inefficient to produce. 
A notable early practical application of STR can be found in \citet{goodfellow2013multi}.
Alternatively, \citet{Jaderberg_STR_2014} treated STR as a word-level classification problem in which each instance was assigned to a 90K-sized dictionary. 
Treating words as varying length sequences as opposed to an explicit character classification has led to large improvements in STR tasks due to the introduction of contextual language information from other characters in the sequence.
Recurrent Neural Networks were used in \cite{shi2016, Wang_STR_2011, shi2017, STAR-Net, Cheng_STR_2017} to encode visual feature maps into one-dimensional sequences.
Connectionist Temporal Classification (CTC) layers were then used to decoder sequences and generate predictions.

\paragraph{Irregular text recognition} Irregular text recognition requires models to handle spatial distortions, noise and large variations in text backgrounds.
\citet{shi2016} use a Spatial Transformer Network in an attempt to mitigate distortion effects and convert irregular text to regular shape.
\citet{Yang_2017} proposed a 2d attention framework for selecting local features, using character-level annotations and a multi-task learning strategy to produce better visual features.
\citet{Li_STR_2019} proposed an LSTM encoder-decoder framework with a 2d attention module, enabling distortions and irregularities to be handled without the need for rectification or character level annotations.
\citet{Baek_2019_ICCV} proposed a common framework for training and evaluation, noting that many previous STR models were trained and evaluated with different strategies, hindering comparisons.


\paragraph{Vision-Language Semantics}
Semantic information has been recently incorporated in other tasks \cite{li2020oscar, singh2019towards}. 
\citet{li2020oscar} proposed using object tags detected in images as anchor points to ease the learning of vision-language alignments in large-scale pre-training of cross-modal representations on image-text pairs.
The semantic information from objects tags significantly improves cross-modal understanding and helps learn generic image-text representations for vision-language understanding and generation tasks.
Our model is inspired by the idea of using semantic information (i.e. object tags) to assist scene text recognition, especially for noisy and hard cases.

\section{Model}

\begin{figure}[t]
\centering
\begin{tabular}{c}
\includegraphics[width=0.95\textwidth]{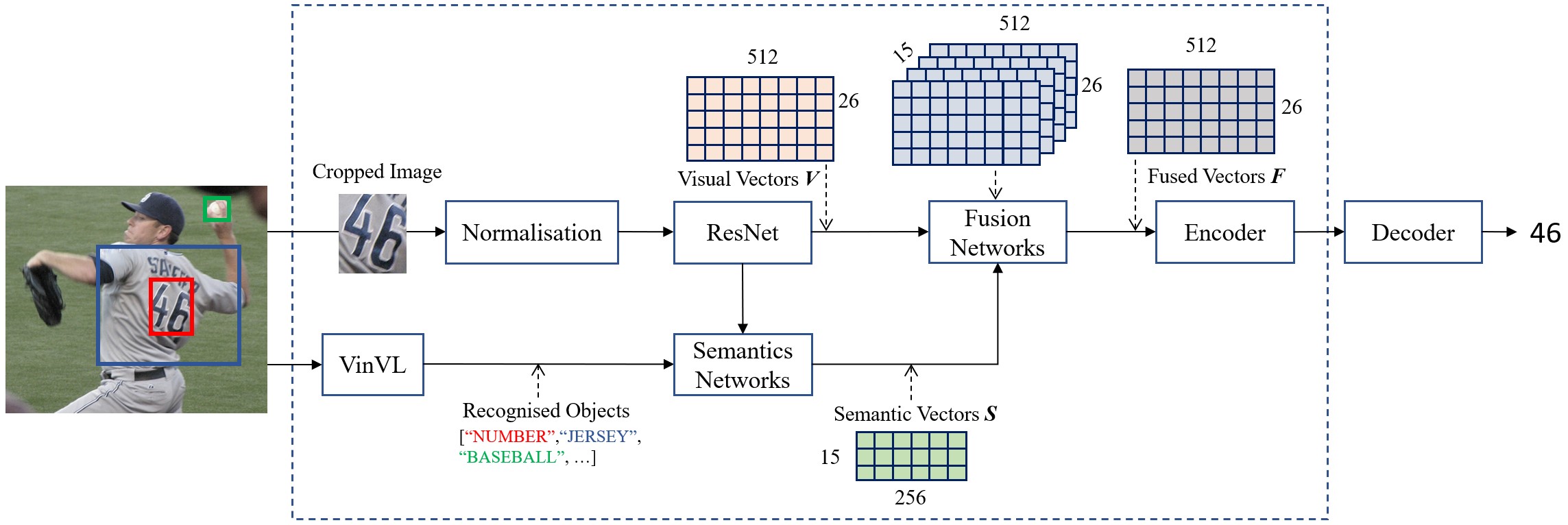} \\
\end{tabular}
\caption{Overview of the modular design of SemanticSTR. Visual and object features are processed separately and then combined within a fusion network before being passed to a transformer \cite{transformer} encoder/decoder structure.}
\label{fig:model}
\end{figure}


In this section we introduce the design and implementation of our semantic STR model (SemanticSTR).
We first introduce the main stages of our STR pipeline (Figure \ref{fig:model}) and subsequently discuss the generation and utilisation of semantic information in SemanticSTR.

\subsection{STR Pipeline}



\textbf{Normalisation}
Word images are first processed by an image normalisation module which is used to transform a cropped text bounding box image $X$ into a normalised image $X'$.
We use the thin-plate spline (TPS) variation of a spatial transformation network originally proposed by \cite{shi2016} and used more recently by \cite{Baek_2019_ICCV, scatter}.
This normalisation essentially allows for subsequent modules to focus less on distortion correction and instead treat inputs as more regular in shape.

\noindent\textbf{Feature Extraction}
Following \cite{Baek_2019_ICCV} we use the ResNet module used in \cite{Cheng_STR_2017} to generate column feature maps conditioned on the normalised image $X'$.
Using pooling $X'$ is split into 26 vertical columns of size 512 creating a sequence of length 26.
The feature map sequence $V = \{v_1, v_2, \cdots, v_{26} \}$ is given as output and corresponds to distinguishable receptive fields used to predict characters.

\noindent\textbf{Semantics}
We deviate from the standard STR architecture by introducing a module which processes information from the greater image scene as semantics.
We use an object detection network to generate object bounding boxes and process them based on their relative position to the text.
We take the natural language tag of each object and use it to map each detected object to a 256 dimensional embedding space, we only consider the 15 most relevant objects to produce a semantic vector $S$ of shape 15x256.
Semantic sequences include padding which gives the model the option to ignore object information by attending exclusively to padding elements (zero vectors).
The semantic vector $S$ is passed to the fusion module which fuses the semantic information into the traditional STR pipeline.
\color{black}

\noindent\textbf{Fusion}
The fusion module takes the semantic vector $S$ and feature map $V$ as input and fuses them into a combined representation.
First $S$ and $V$ are projected to a common space of shape $V_{length}$ x $S_{length}$ x ($V_{feature size}$ + $S_{feature size}$): 26x15x(512+256).
We then use attention-weighted summation and linear mapping to preserve only the most relevant object information producing a sequence $F$ of shape 26x512, specific details on the internal structure of the fusion network are provided in subsection \ref{sec:semantics}.
\color{black}

\noindent\textbf{Encoder-Decoder}
We use a encoder-decoder transformer \cite{transformer} architecture to process features and generate character predictions.
The encoder takes $F$ from the fusion module and remaps it, introducing inter-sequence contextual information.
$F$ has length 26 but most words will be of shorter length, e.g. the example in Figure \ref{fig:model} only contains the characters "4" and "6", ideally the encoder should produce a new sequence $F'$ with $f_0$ storing information for the first character in the sequence and $f_1$ storing information for the second character etc.
The decoder iterates over $F'$ generating predictions at each step.
At step $n$ the decoder looks at the features at $f'_n$ and also all previous predicted characters from steps \{$0$, ..., $n-1$\} to calculate the $nth$ character. 
The decoder returns an end-of-sequence token when it believes the text sequence is complete.

\subsection{Semantics}
\label{sec:semantics}
\begin{figure}[H]
\vspace{-0.1cm}
\centering
\begin{tabular}{cc}
\includegraphics[height=2.2cm]{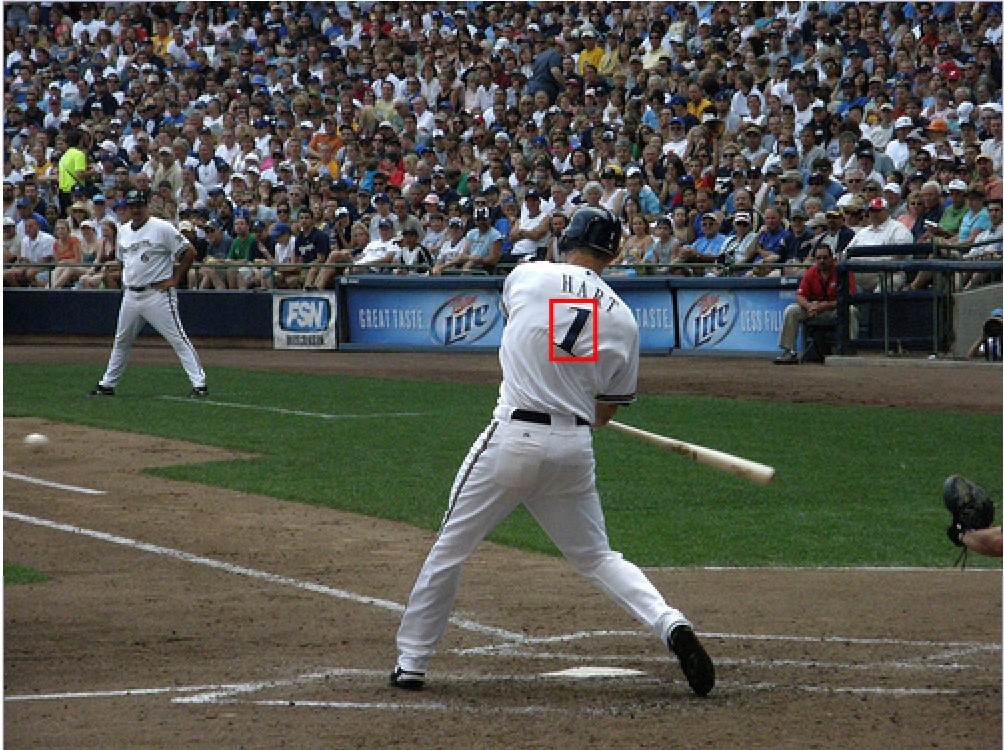} &
\includegraphics[height=2.2cm]{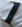} \\
(a) Scene Image & (b) Cropped Image
\end{tabular}
\caption{Comparison of full scene image vs cropped word image}
\vspace{-0.3cm}
\label{fig:scene_vs_crop}
\end{figure}

Our primary contribution is to introduce scene semantic information into STR model design.
We use an object detector to generate a semantic vector that stores object information about the scene.
For example, in Figure \ref{fig:scene_vs_crop}, just considering the cropped image (b) it can be difficult to distinguish between the characters "l", "1", and "I".
With access to the scene image (a), however, we can reason that numbers are typically present on the back of sports jerseys and predict that this is the digit "1".

\begin{figure}[H]
\centering
\begin{tabular}{ccc}
\includegraphics[height=2.3cm]{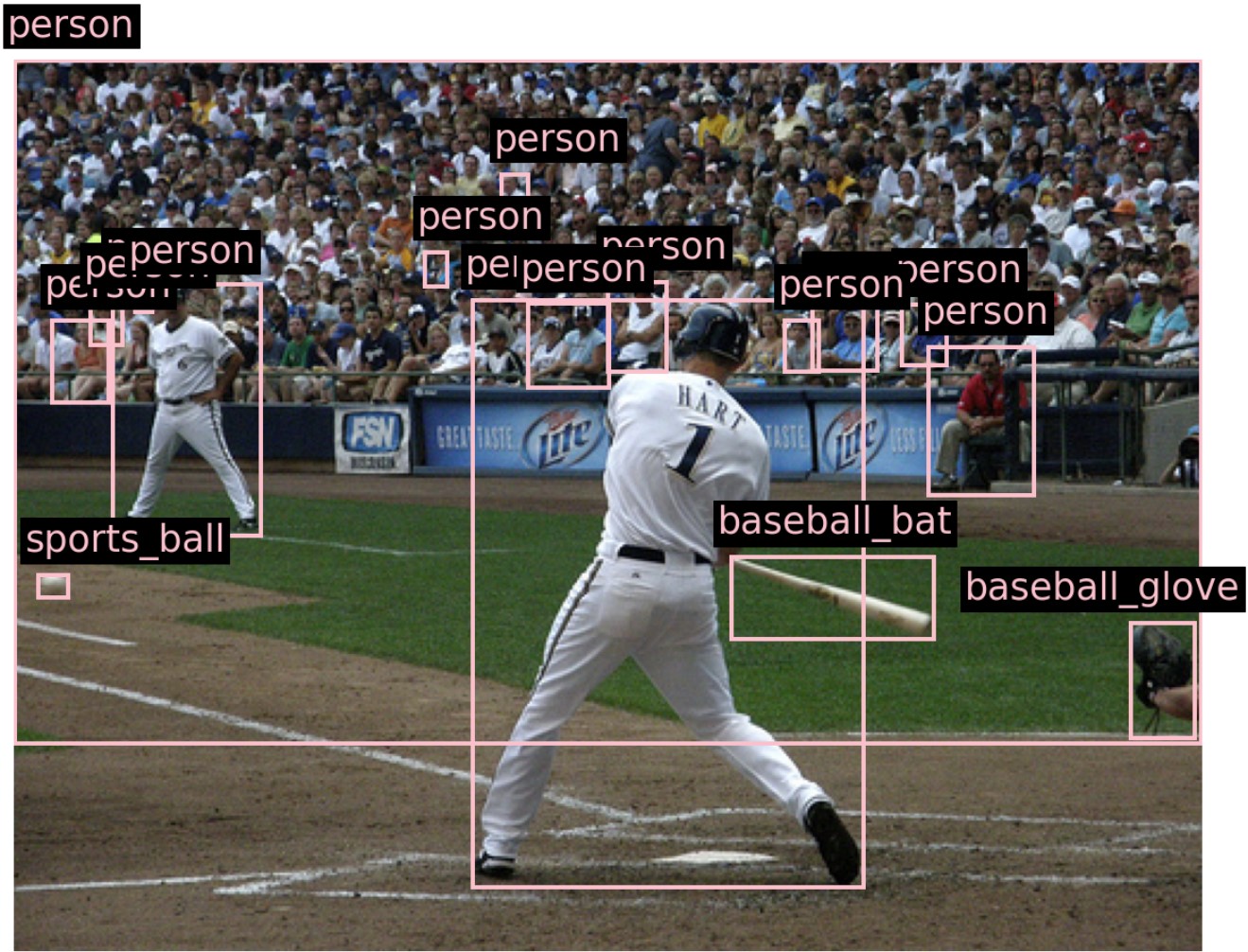} &
\includegraphics[height=2.175cm]{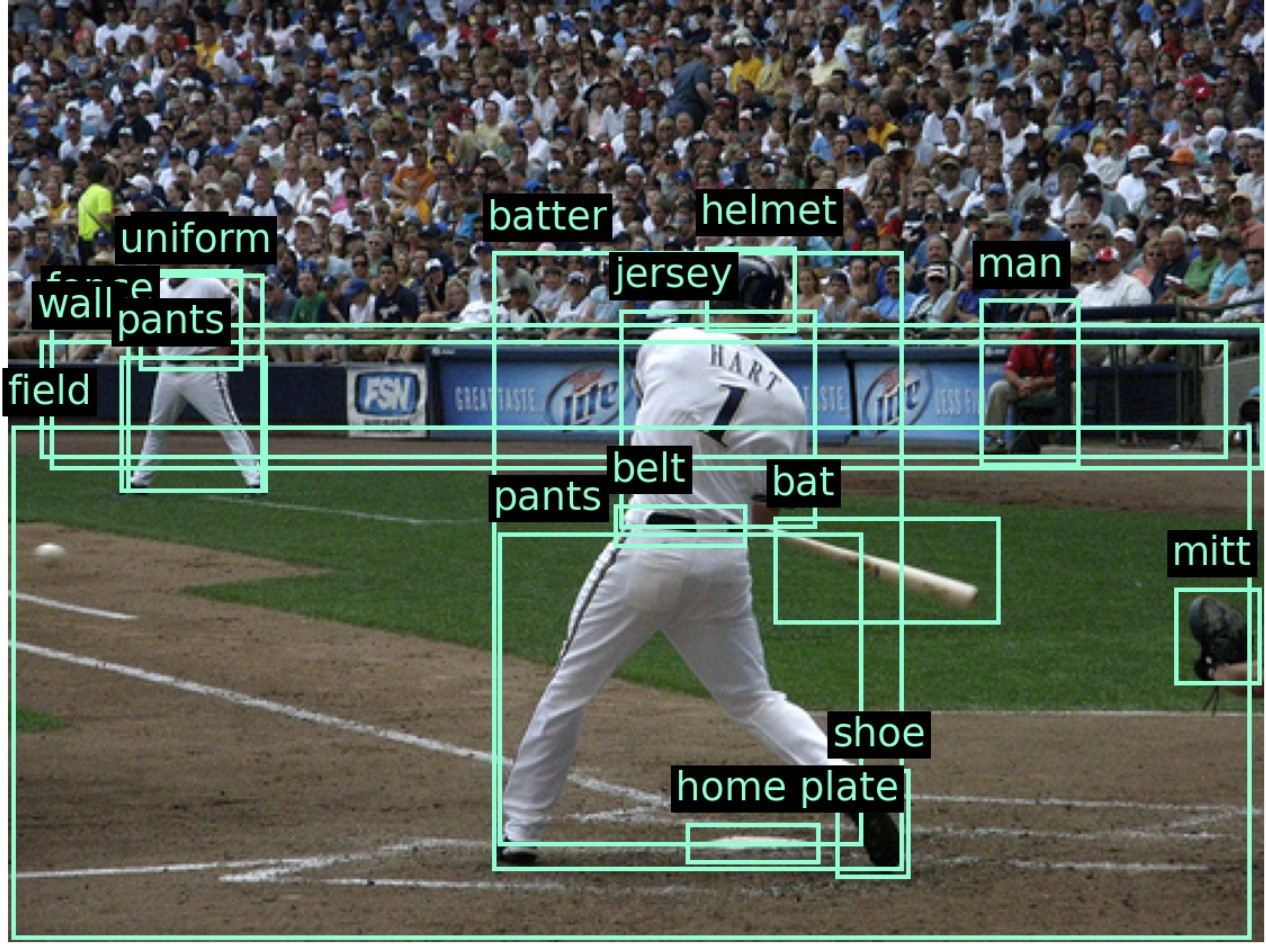} &
\includegraphics[height=2.3cm]{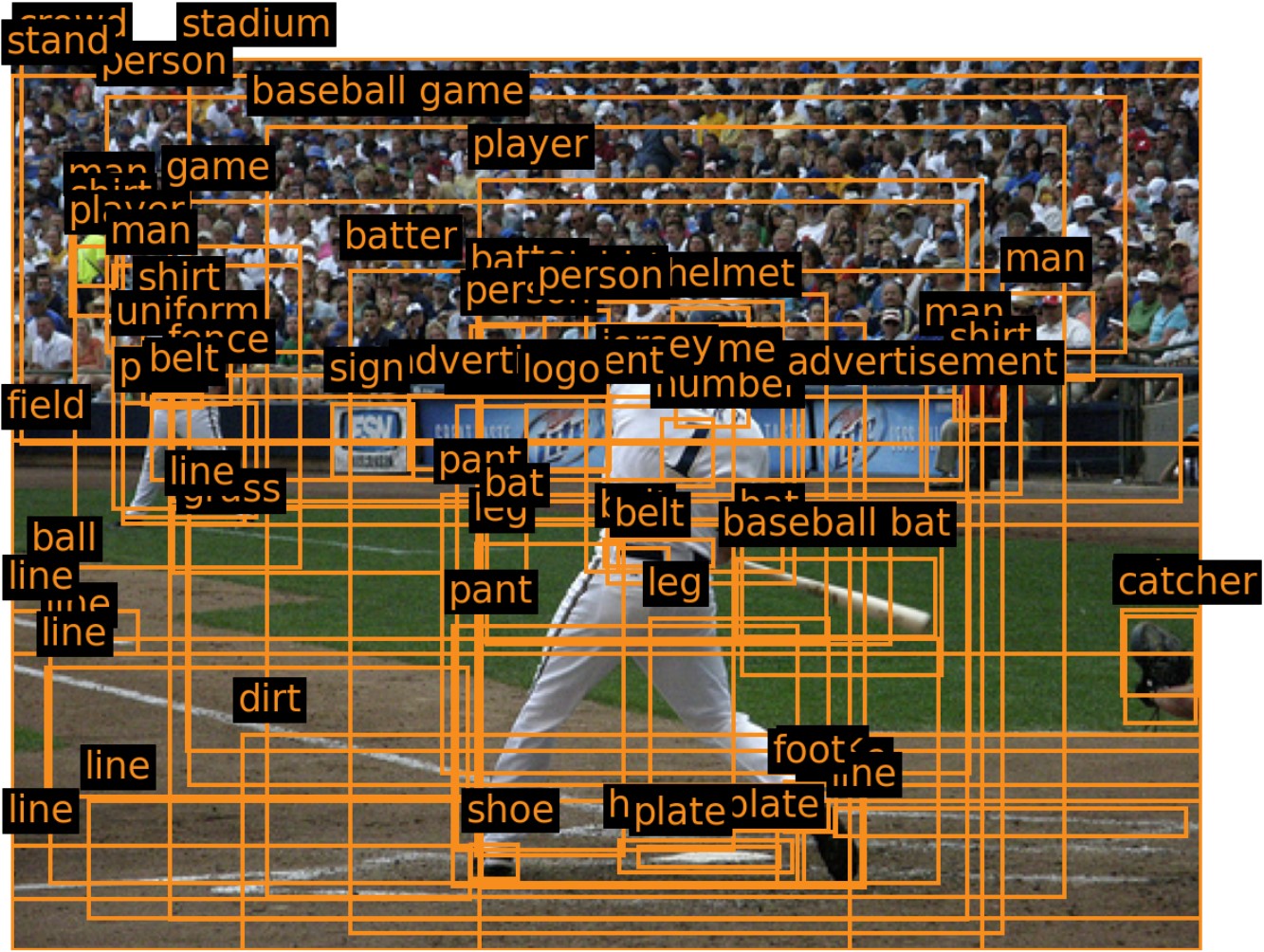} \\
(a) MS-COCO & (b) Visual Genome & (c) VinVL
\end{tabular}
\caption{Visualisation of generated object bounding boxes for an object detector trained on the MS-COCO (a) \cite{ms_coco}, Visual Genome (b) \cite{visual_genome}, and VinVL (c) \cite{zhang2021vinvl} datasets.}
\label{fig:source_comparison}
\end{figure}

We initially experimented with three different object detectors, each trained on a different established object dataset.
Figure \ref{fig:source_comparison} shows a comparison of MS-COCO \cite{ms_coco}, Visual Genome \cite{visual_genome}, and VinVL \cite{zhang2021vinvl} trained object detectors.
We established that the range and quantity of VinVL generated objects was far richer for our task than MS-COCO and Visual Genome and so use its detections for our final model.

\begin{figure}[H]
\centering
\begin{tabular}{c}
\includegraphics[width=0.85\textwidth]{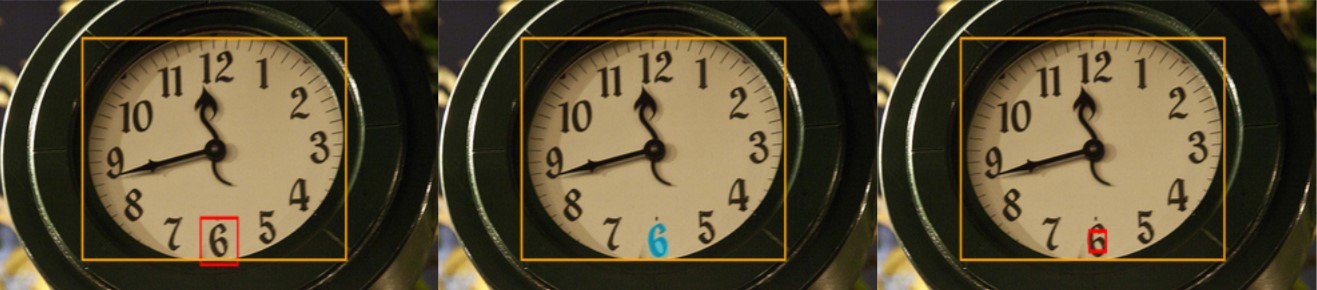} \\
    \begin{tabular}{p{0.3\textwidth}p{0.3\textwidth}p{0.3\textwidth}}
    \hfil A) Original Box & \hfil B) Mask Area & \hfil C) Scaled Box \\
    \end{tabular} \\
\end{tabular}
\vspace{-0.3cm}
\caption{Scaling annotation by its bounding box and segmentation mask area}
\vspace{-0.3cm}
\label{fig:overlap_rescale}
\end{figure}

Using generated object bounding boxes and class labels we explore two methodologies for assigning object representation to text samples:




\textbf{Overlap Semantics}
Objects are assigned as overlap semantics based on their relative position with respect to the text instance.
We scale text bounding boxes by their segmentation mask area in proportion to their bounding box area, then an object is assigned if it fully encompasses the scaled text box.
A visual example of scaling text bounding boxes can be seen in Figure \ref{fig:overlap_rescale}.
The scale calculations are defined as follows:
$box_{\text{old}} = [x, y, w, h]$, 
$s = \frac{\text{mask}_{\text{area}}}{\text{box}_{\text{area}}}$,
\begin{equation}
\label{eq:box_rescale}
    box_{new} = [x + (1 - s)\frac{w}{2}, y + (1-s)\frac{h}{2}, sw, sh]
\end{equation}
All objects in the scene that completely encompass the scaled text bounding box are assigned to the text instance.
An object encompasses the text if all four corners of the texts bounding box are within the border of the objects bounding box.

\textbf{Scene Semantics}
Instead of selectively gathering the object semantics by relative positions, a scene vector includes all objects detected in the scene, giving the model access to all semantic information.
In order to distil some form of positional encoding into the scene vectors we calculate the insertion area of both bounding boxes and divide this by the union of the boxes areas (IoU scores). These scores are used to scale the objects information, with a minimum score of 0.1 set to prevent objects being completely ignored. \\
\color{black}

\noindent
To process semantics vectors, we pass object tags through a pretrained BERT \cite{devlin-etal-2019-bert} model to generate \emph{768} dimensional representations.
BERT is a transformer encoder pretrained on large semi-supervised language tasks.
Due to the high number of objects classes presented by VinVL it is important for our model to interpret similarities between classes.
For example, the labels "CAR" and "TRUCK" should map to similar embedding representations and should be distant from "SANDWICH" for example.

Some object tags are more relevant to text than others.
Consider Figure \ref{fig:scene_vs_crop} again, the object tag "NUMBER" is more useful than "PERSON" in this situation for contextualising a prediction of the digit "1".
Different samples can also have a different number of objects associated with them, to address both of these issues we compute normalised relevancy scores.
Given the visual feature sequence $V$ from the feature extraction model and a sequence of object embeddings $S$, we calculate separate object relevancy scores for each of the visual feature elements \{$v_0$, ..., $v_{26}$\}.
For a visual feature sequence of length 26 and object sequence of length 15 we calculate 26x15 relevancy scores.
These scores are calculated by concatenating features from a visual element $v_i$ and a singular object embedding $s_j$ and then passing these through a multi-layer perception (MLP).
We then normalise the computed scores by applying softmax over the object sequence at each $v_i$, thus calculating normalised weights for each objects relevance.
We then sum along the object sequence and produce a sequence of vectors $F$ which contains fused information from both $V$ and $S$:
\begin{equation}
    f_i = v_i + \sum_{j}\text{softmax}(\text{MLP}(\text{concat}(v_i, s_j)))
\end{equation}
A residual connection is applied to fuse visual and normalised object features.
This fusion module can be placed at various stages on the STR pipeline.
We experiment with placements pre-encoder, pre-decoder and post-decoder.
We also gather results from 2 additional fusion strategies:
\begin{enumerate}
\item Inserting an additional multi-head attention block in the decoder which uses the target sequence as the query and the semantic vector as the key and value.
\item Initialising the decoder by replacing the standard [CLS] token with a semantic vector.
\end{enumerate}
We ultimately found these two methods achieve poor performance compared to the approach implemented in our final model but record their performance for comparison.
The results of all fusion approaches are reported in Table \ref{table:fusion_results}.

\section{Experiments}

\subsection{Datasets}

Three different image-text datasets are used to train and evaluate our models: 

\textbf{Synthetic Dataset}
Following  \citet{Baek_2019_ICCV}, synthetic datasets are used to pre-train model architectures before fine-tuning on `real' samples. 
Specifically, we use a union of the MJSynth (MJ) \cite{MJ_Synth_Jaderberg} and SynthText (ST) \cite{ST_Synth_Gupta} datasets.

\textbf{Cropped Dataset}
Conventional STR models use "cropped" word samples for training, that is, images that tightly encompass a single word to be recognized. 
Here we only use cropped datasets during synthetic pre-training, specifically we use the regular text datasets
IC13 \cite{IC13}, IIIT5K \cite{Mishra_BMVC}, SVT \cite{Wang_STR_2011} and irregular text dataset IC15 \cite{IC15}.

\textbf{Full Scene Dataset}
"Full Scene" datasets source images from other established tasks such as object detection and provide word level annotations for a subset of images.
In this work we use COCOText \cite{veit2016cocotext} and TextOCR \cite{singh2021textocr} to train the proposed STR model with semantics.
COCOText provides annotations for the MS-COCO 2014 \cite{ms_coco} train image set and gives useful instance information such as a boolean value for the legibility of a word and the typeface source (machine printed or handwritten).
TextOCR provides annotations for a subset of the OpenImages \cite{OpenImages} training set and also provides legibility information.
More details for COCOText and TextOCR can be found in \citet{ms_coco} and \citet{singh2021textocr} respectfully.


\subsection{Training}
We follow the common practice established in \cite{Baek_2019_ICCV} and pre-train our transformer architecture for 300,000 iterations on the union of the MJ and ST synthetic datasets.
We use a batch size of 192 and evaluate the model every 2000 iterations on the union of the training sets of IC13, IC15, IIIT, and SVT, we save the highest performing checkpoint as our final pre-train checkpoint.
Note that these are `cropped word' datasets that do not provide complete scenes, so we are only training our underling backbone and transformer architecture without semantic fusion.
Our semanticSTR model is finetuned on COCOText and TextOCR respectively.
We evaluate our model every 200 and 2000 iterations on COCOText and TextOCR, and early stop based on the performance on validation datasets.
Table \ref{table:full_results} reports the median performance checkpoint from multiple runs. The transformer model contains 6 encoding layers and 6 decoding layers of hidden size 512.
Cross entropy loss is used with the AdamW optimiser \cite{AdamW} and a learning rate $1e^{-4}$ scheduled to decrease on loss plateaus.
All results are recorded as word-level correct predictions.
\color{black}

\subsection{Results}

\begin{table}[!t]
\centering
\resizebox{0.9\linewidth}{!}{
\begin{tabular}{c|c|cccc}
 \toprule
 \multirow{2}{*}{Model}& \multirow{2}{*}{Pretrain}& Train & \multicolumn{2}{c}{\texttt{COCOText}}& \texttt{TextOCR}\\
 \cline{3-6}
     &  & Test & \texttt{COCOText} & \texttt{TextOCR} & \texttt{TextOCR}\\
 \hline
    TPS-ResNet-BiLSTM-Attn \cite{Baek_2019_ICCV} & - && 50.16 & 36.41 & 65.65 \\
    TransformerSTR (ours) & - && 47.77  &  34.31 & 70.89 \\
    SemanticSTR (ours) & - && \textbf{50.54} &  \textbf{38.28} & \textbf{74.31} \\
 \hline
    CRNN \cite{shi2016} & \texttt{MJ+ST} && 49.00 &  43.07 & 58.61 \\
    Rosetta \cite{Rosetta} & \texttt{MJ+ST} && 49.66 &  43.16 & 60.85 \\
    STAR-Net \cite{STAR-Net} & \texttt{MJ+ST} && 53.56 &  48.23 & 66.84 \\
    TPS-ResNet-BiLSTM-Attn \cite{Baek_2019_ICCV} & \texttt{MJ+ST} && 59.32 &  50.37 & 69.49 \\
    TransformerSTR (ours) & \texttt{MJ+ST} && 63.71 &  55.05 & 76.30 \\
    SemanticSTR (ours) & \texttt{MJ+ST} && \textbf{64.33} &  \textbf{56.7} & \textbf{76.71} \\
\bottomrule
\end{tabular}
}
\caption{
Comparison of STR models on COCOText and TextOCR, with and without synthetic pretraining (MJ+ST).
The CRNN \cite{shi2016}, Rosetta \cite{Rosetta}, STAR-Net \cite{STAR-Net} and TPS-ResNet-BiLSTM-Attn \cite{Baek_2019_ICCV} are the benchmark end-to-end neural models for scene text recognition.
} 
\label{table:full_results}
\end{table}
\begin{table}[!t]
\centering
\footnotesize
\resizebox{0.8\linewidth}{!}{
\begin{tabular}{ c|ccc }
 \toprule
 Fusion Strategy & Semantic Vector & COCOText & TransformerSTR Failure Corrections*\\
 \hline
  TransformerSTR & None & 63.71 & 0 \\
  \hline
   Pre-Encoder MLP & \begin{tabular}{@{}c@{}}Overlap \\ Scene\end{tabular} & \begin{tabular}{@{}c@{}}\textbf{64.33} \\ 63.66\end{tabular} & \begin{tabular}{@{}c@{}}\textbf{9.78} \\ 8.47 \end{tabular} \\
   Pre-Decoder MLP & \begin{tabular}{@{}c@{}}Overlap \\ Scene\end{tabular} & \begin{tabular}{@{}c@{}}63.83 \\ 63.57\end{tabular} & \begin{tabular}{@{}c@{}}8.63 \\ 8.01 \end{tabular} \\
   Post-Decoder MLP & \begin{tabular}{@{}c@{}}Overlap \\ Scene\end{tabular} & \begin{tabular}{@{}c@{}}60.06 \\ 59.89\end{tabular} & \begin{tabular}{@{}c@{}}4.43 \\  4.26\end{tabular} \\

  \hline
  Object CLS Token & \begin{tabular}{@{}c@{}}Overlap \\Scene\end{tabular} & \begin{tabular}{@{}c@{}}63.72 \\ 63.61\end{tabular} & \begin{tabular}{@{}c@{}}8.55 \\ 8.25\end{tabular} \\
  Multihead Pre-Memory MLP & \begin{tabular}{@{}c@{}}Overlap \\Scene\end{tabular} & \begin{tabular}{@{}c@{}}63.60 \\ 63.59\end{tabular} & \begin{tabular}{@{}c@{}}8.63 \\ 7.28\end{tabular} \\
 \bottomrule
\end{tabular}
}
\caption{Modality Fusion Results. 
Strong performance on TransformerSTR failure corrections suggests semantic information can improve the models robustness to hard text samples. \\
*TransformerSTR failure corrections refers to samples which our baseline transformer model failed to recognize, we then report the accuracies our other models achieved on this set of "hard" samples.
}
\label{table:fusion_results}
\end{table}
Table \ref{table:full_results} compares SemanticSTR to our TransformerSTR baseline model and other high performing STR models on COCOText and TextOCR.
TransformerSTR is our transformer architecture without the inclusion of semantic information and allows us to compare the direct performance effect of our semantic modules.
Firstly, the transformer-based encoder/decoder structure mostly improves the performance of the STR pipeline on both datasets.
SemanticSTR is able to boost performance over TransformerSTR in all experiments, however we find this difference is more marginal after significant synthetic pretraining.
We hypothesise that this is due to the inability of the models to fluidly adapt to semantic inputs after being heavily pretrained with non-semantic samples.
Therefore semantic representations may be useful in situations when robust pretraining is not available.
\color{black}
Furthermore, despite relatively similar total accuracy scores with pretraining, SemanticSTR deviates greatly from TransformerSTR in terms of the samples it can consistently recognise.
Analysing performance on COCOText we record the set of samples that the baseline model is able to correctly identify and the set of samples incorrectly identified.
We then analyse the performance of SemanticSTR on these subsets.
Our results show that on TransformerSTR failure cases our semantic model is able to correct 486/4969 samples, however on the baseline correct set our semantic model incorrectly identifies 398/8764 samples, resulting in a net 88 sample improvement. Section \ref{results_analysis} explores these sample groups in detail.

\paragraph{Fusion Strategies}
We experiment with combining visual and semantic features at different stages in the STR pipeline and record our results in Table \ref{table:fusion_results}.
Semantic fusion is applied pre-encoder, pre-decoder, or post-decoder of the transformer architecture.
We also attempt fusion during the multihead attention component of the decoder block and finally experiment with replacing the [CLS] token from the start of the decoder target sequence with the semantic vector.
This [CLS] replacement effectively initialises the transformer decoder, which has previously proven to be effective in RNNs \cite{rnn_init}.
We evaluate our fusion strategies using total word-level accuracy and on TransformerSTR failure cases which are exclusively samples TransformerSTR fails on.
\color{black}
We identify strong correlations between TransformerSTR failure cases and hard text samples, and so use these samples to report hard text performance.
Table \ref{table:fusion_results} shows pre-encoder fusion to outperform all other fusion strategies.

    
    
    
    
    
    
    

\subsection{Analysis} \label{results_analysis}

\begin{figure}[t]
\centering
\begin{tabular}{c}
\includegraphics[width=0.95\textwidth]{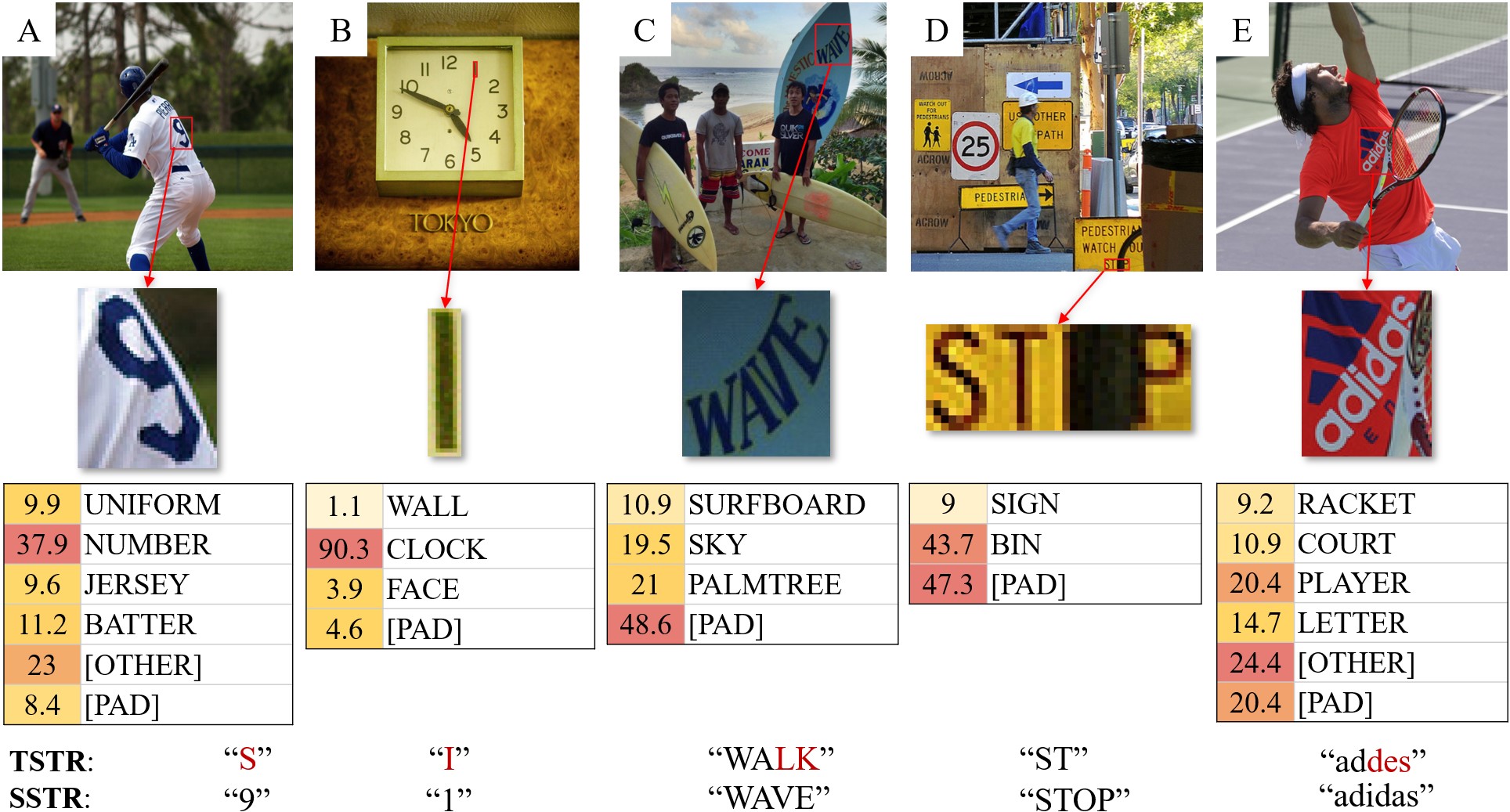}
\end{tabular}
\caption{SemanticSTR corrections. In (B) SemanticSTR attends to the "CLOCK" object tag to correctly identify the digit `1' which the baseline model predicts as `I'.
(D) contains significant character obstruction that the baseline model is unable to handle, SemanticSTR is able to correctly predict "STOP" in this situation.
Here (C) and (D) both show situations where the model attends heavily to the padding and partially to the objects, suggesting that semantics are less beneficial in these cases.
This reinforces that the model is able to selectively attend to objects when needed, but also can learn to ignore objects.}
\label{fig:semantic_corrections}
\end{figure}

\paragraph{Transformer vs RNN}
The results displayed in table \ref{table:full_results} show that in most of our experiments our transformer models outperform their RNN counter parts.
We do however find that TransformerSTR fails to outperform the LSTM model on COCOText without synthetic pretraining.
SemanticSTR is able to achieve higher scores with no pretraining, leading us to believe that using semantics can mitigate the negative effects of small training sets for transformer models.
This is most notable on TextOCR where SemanticSTR is able to significantly outperform TransformerSTR and the LSTM model by 3.42 and 8.66 percentage points respectfully.
We note however that the difference in performance fluctuates depending on the training data, more specifically the quantity of training data.
Training on smaller datasets, such as COCOText, can diminish the potential improvements a transformer structure can offer due to lack of samples.
\color{black}

\begin{figure}[!t]
\centering
\begin{tabular}{c}
\includegraphics[width=0.95\textwidth]{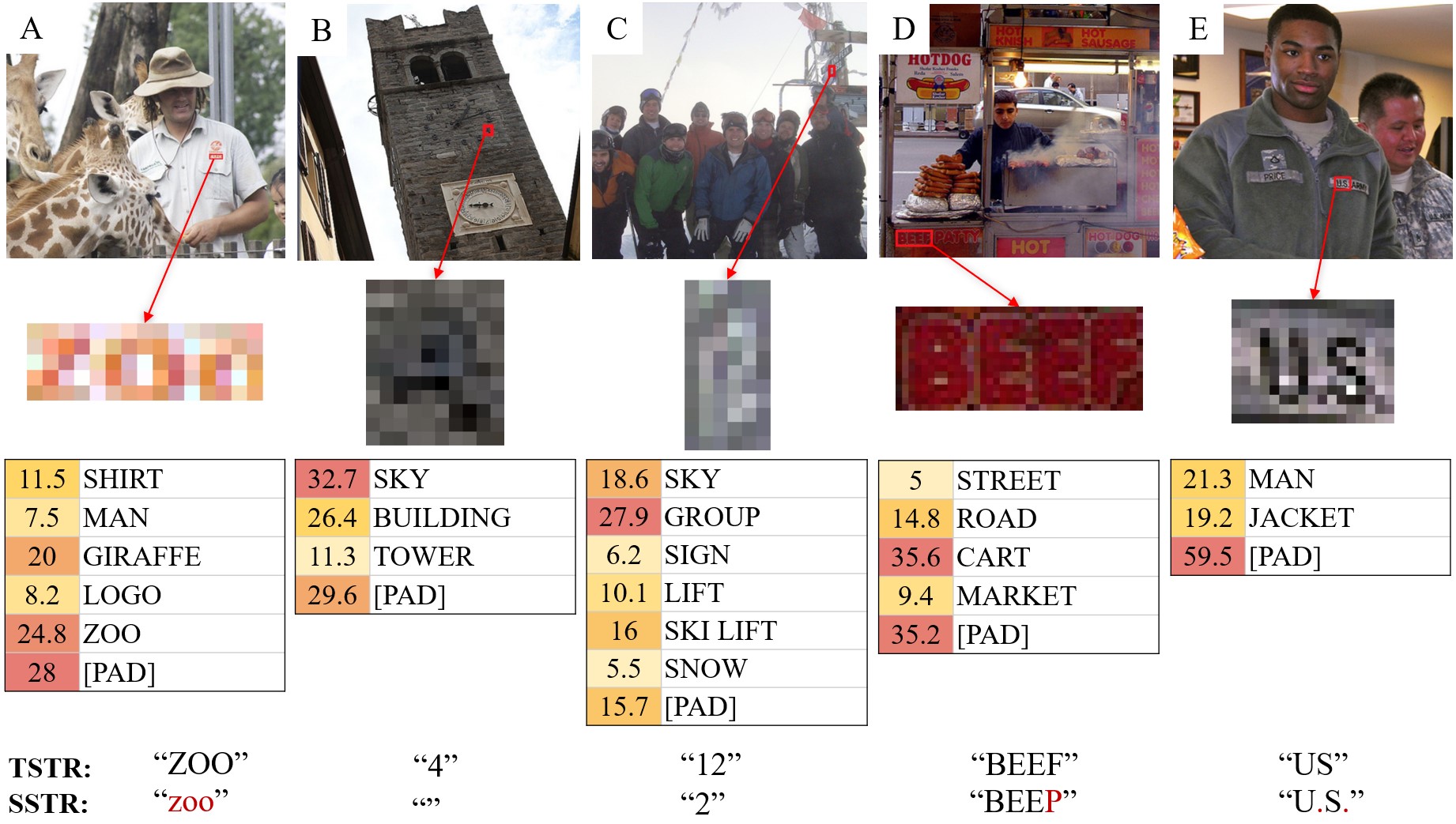}
\end{tabular}
\caption{SemanticSTR failures. A common factor of semantic failure cases is extremely low resolution images.
This reduction in capacity results in errors being made on challenging examples such as (B) and (C).
Another contributing factor to SemanticSTR failures cases is label noise, (E) shows an example of a instance that SemanticSTR was able to correctly predict but due to mislabelling credit is wrong attributed to the baseline prediction.}
\label{fig:semantic_failures}
\end{figure}

\paragraph{TransformerSTR vs SemanticSTR}
To qualitatively compare SemanticSTR and TransformerSTR, we show samples that only one of these models was able to successfully interpret.
Figure \ref{fig:semantic_corrections} shows COCOText samples only SemanticSTR correctly predicts, figure \ref{fig:semantic_failures} shows samples only TransformerSTR predicts correctly.
Both figures show the full scene image, word cropped image and the calculated object relevancy scores.
In cases where large numbers of objects are assigned we have included an [OTHER] tag which is a summation of the attention scores of the less influential objects.
We report that SemanticSTR is able to outperform TransformerSTR on all experiments, however the performance gain varies depending on the task. See figure captions for details.
\color{black}

\paragraph{SemanticSTR Corrections} We observe that SemanticSTR outperforms TransformerSTR in situations where text is partially obscured, difficult to identify due to common character structure (e.g. `I', `l' and `1'), or contains significant visual distortion, i.e. hard text samples.
Figure \ref{fig:semantic_corrections} shows five examples of such corrections.
\color{black}

\paragraph{SemanticSTR Failures}
We identify that the most common cause of SemanticSTR failures is low image resolution.
We hypothesise that this may be caused by a reduction in the ability to retain intricate visual information because of the addition of semantics to the same feature space.
Low resolution images also introduce potential labelling noise due to the human annotators inability to distinguish complete characters.
(E) in figure \ref{fig:semantic_failures} is an example where the ground truth label has ignored the punctuation points in the sequence "U.S." resulting in SemanticSTR's prediction being rendered incorrect.
\color{black}



\section{Conclusion}
STR remains an unsolved problem.
The lack of awareness of the greater image scene can make predicting partially obscured and heavily distorted sequences challenging.
We propose a multimodal approach generating semantic information using object tags.
SemanticSTR is able to selectively attend to scene objects to contextualise predictions, correcting a significant number of baseline non-semantic model errors.
Our semantic approach achieves higher performance on full scene STR tasks with and without synthetic pretraining.
We show the existence of relationships between objects and text within a common scene, and that utilising these connections can improve the performance of STR models, particularly on hard text.

\bibliography{egbib}

\begin{thebibliography}{39}
\providecommand{\natexlab}[1]{#1}
\providecommand{\url}[1]{\texttt{#1}}
\expandafter\ifx\csname urlstyle\endcsname\relax
  \providecommand{\doi}[1]{doi: #1}\else
  \providecommand{\doi}{doi: \begingroup \urlstyle{rm}\Url}\fi

\bibitem[Baek et~al.(2019)Baek, Kim, Lee, Park, Han, Yun, Oh, and
  Lee]{Baek_2019_ICCV}
Jeonghun Baek, Geewook Kim, Junyeop Lee, Sungrae Park, Dongyoon Han, Sangdoo
  Yun, Seong~Joon Oh, and Hwalsuk Lee.
\newblock What is wrong with scene text recognition model comparisons? dataset
  and model analysis.
\newblock In \emph{Proceedings of the IEEE/CVF International Conference on
  Computer Vision (ICCV)}, October 2019.

\bibitem[Bartz et~al.(2017)Bartz, Yang, and Meinel]{STN-OCR}
Christian Bartz, Haojin Yang, and Christoph Meinel.
\newblock {STN-OCR:} {A} single neural network for text detection and text
  recognition.
\newblock \emph{CoRR}, abs/1707.08831, 2017.
\newblock URL \url{http://arxiv.org/abs/1707.08831}.

\bibitem[Borisyuk et~al.(2018)Borisyuk, Gordo, and Sivakumar]{Rosetta}
Fedor Borisyuk, Albert Gordo, and V.~Sivakumar.
\newblock Rosetta: Large scale system for text detection and recognition in
  images.
\newblock \emph{Proceedings of the 24th ACM SIGKDD International Conference on
  Knowledge Discovery \& Data Mining}, 2018.

\bibitem[Cheng et~al.(2017)Cheng, Bai, Xu, Zheng, Pu, and Zhou]{Cheng_STR_2017}
Zhanzhan Cheng, Fan Bai, Yunlu Xu, Gang Zheng, Shiliang Pu, and Shuigeng Zhou.
\newblock Focusing attention: Towards accurate text recognition in natural
  images.
\newblock \emph{2017 IEEE International Conference on Computer Vision (ICCV)},
  pages 5086--5094, 2017.

\bibitem[Devlin et~al.(2019)Devlin, Chang, Lee, and
  Toutanova]{devlin-etal-2019-bert}
Jacob Devlin, Ming-Wei Chang, Kenton Lee, and Kristina Toutanova.
\newblock {BERT}: Pre-training of deep bidirectional transformers for language
  understanding.
\newblock In \emph{Proceedings of the 2019 Conference of the North {A}merican
  Chapter of the Association for Computational Linguistics: Human Language
  Technologies, Volume 1 (Long and Short Papers)}, pages 4171--4186,
  Minneapolis, Minnesota, June 2019. Association for Computational Linguistics.
\newblock \doi{10.18653/v1/N19-1423}.
\newblock URL \url{https://www.aclweb.org/anthology/N19-1423}.

\bibitem[Goodfellow et~al.(2013)Goodfellow, Bulatov, Ibarz, Arnoud, and
  Shet]{goodfellow2013multi}
Ian~J Goodfellow, Yaroslav Bulatov, Julian Ibarz, Sacha Arnoud, and Vinay Shet.
\newblock Multi-digit number recognition from street view imagery using deep
  convolutional neural networks.
\newblock \emph{arXiv preprint arXiv:1312.6082}, 2013.

\bibitem[Gupta et~al.(2016)Gupta, Vedaldi, and Zisserman]{ST_Synth_Gupta}
Ankush Gupta, Andrea Vedaldi, and Andrew Zisserman.
\newblock Synthetic data for text localisation in natural images.
\newblock In \emph{IEEE Conference on Computer Vision and Pattern Recognition},
  2016.

\bibitem[Islam et~al.(2017)Islam, Islam, and Noor]{Islam_ocr_survey}
Noman Islam, Zeeshan Islam, and Nazia Noor.
\newblock A survey on optical character recognition system.
\newblock \emph{CoRR}, abs/1710.05703, 2017.
\newblock URL \url{http://arxiv.org/abs/1710.05703}.

\bibitem[Jaderberg et~al.(2014{\natexlab{a}})Jaderberg, Simonyan, Vedaldi, and
  Zisserman]{Jaderberg_STR_2014}
Max Jaderberg, K.~Simonyan, A.~Vedaldi, and Andrew Zisserman.
\newblock Synthetic data and artificial neural networks for natural scene text
  recognition.
\newblock \emph{ArXiv}, abs/1406.2227, 2014{\natexlab{a}}.

\bibitem[Jaderberg et~al.(2014{\natexlab{b}})Jaderberg, Simonyan, Vedaldi, and
  Zisserman]{MJ_Synth_Jaderberg}
Max Jaderberg, Karen Simonyan, Andrea Vedaldi, and Andrew Zisserman.
\newblock Synthetic data and artificial neural networks for natural scene text
  recognition.
\newblock In \emph{Workshop on Deep Learning, NIPS}, 2014{\natexlab{b}}.

\bibitem[Karatzas et~al.(2013)Karatzas, Shafait, Uchida, Iwamura, Bigorda,
  Mestre, Mas, Mota, Almazan, and {De Las Heras}]{IC13}
Dimosthenis Karatzas, Faisal Shafait, Seiichi Uchida, Masakazu Iwamura, {Lluis
  Gomez I.} Bigorda, {Sergi Robles} Mestre, Joan Mas, {David Fernandez} Mota,
  {Jon Almazan} Almazan, and {Lluis Pere} {De Las Heras}.
\newblock Icdar 2013 robust reading competition.
\newblock \emph{Proceedings of the International Conference on Document
  Analysis and Recognition, ICDAR}, pages 1484--1493, 2013.
\newblock ISSN 1520-5363.
\newblock \doi{10.1109/ICDAR.2013.221}.
\newblock Copyright: Copyright 2013 Elsevier B.V., All rights reserved.; 12th
  International Conference on Document Analysis and Recognition, ICDAR 2013 ;
  Conference date: 25-08-2013 Through 28-08-2013.

\bibitem[Karatzas et~al.(2015)Karatzas, Gomez-Bigorda, Nicolaou, Ghosh,
  Bagdanov, Iwamura, Matas, Neumann, Chandrasekhar, Lu, Shafait, Uchida, and
  Valveny]{IC15}
Dimosthenis Karatzas, Lluis Gomez-Bigorda, Anguelos Nicolaou, Suman Ghosh,
  Andrew Bagdanov, Masakazu Iwamura, Jiri Matas, Lukas Neumann,
  Vijay~Ramaseshan Chandrasekhar, Shijian Lu, Faisal Shafait, Seiichi Uchida,
  and Ernest Valveny.
\newblock Icdar 2015 competition on robust reading.
\newblock In \emph{Proceedings of the 2015 13th International Conference on
  Document Analysis and Recognition (ICDAR)}, ICDAR '15, page 1156–1160, USA,
  2015. IEEE Computer Society.
\newblock ISBN 9781479918058.
\newblock \doi{10.1109/ICDAR.2015.7333942}.
\newblock URL \url{https://doi.org/10.1109/ICDAR.2015.7333942}.

\bibitem[Krishna et~al.(2016)Krishna, Zhu, Groth, Johnson, Hata, Kravitz, Chen,
  Kalantidis, Li, Shamma, Bernstein, and Fei-Fei]{visual_genome}
R.~Krishna, Yuke Zhu, O.~Groth, J.~Johnson, K.~Hata, J.~Kravitz, Stephanie
  Chen, Yannis Kalantidis, Li-Jia Li, D.~Shamma, Michael~S. Bernstein, and
  Li~Fei-Fei.
\newblock Visual genome: Connecting language and vision using crowdsourced
  dense image annotations.
\newblock \emph{International Journal of Computer Vision}, 123:\penalty0
  32--73, 2016.

\bibitem[Kuznetsova et~al.(2020)Kuznetsova, Rom, Alldrin, Uijlings, Krasin,
  Pont-Tuset, Kamali, Popov, Malloci, Kolesnikov, Duerig, and
  Ferrari]{OpenImages}
Alina Kuznetsova, Hassan Rom, Neil Alldrin, Jasper Uijlings, Ivan Krasin, Jordi
  Pont-Tuset, Shahab Kamali, Stefan Popov, Matteo Malloci, Alexander
  Kolesnikov, Tom Duerig, and Vittorio Ferrari.
\newblock The open images dataset v4: Unified image classification, object
  detection, and visual relationship detection at scale.
\newblock \emph{IJCV}, 2020.

\bibitem[Li et~al.(2019)Li, Wang, Shen, and Zhang]{Li_STR_2019}
Hui Li, Peng Wang, Chunhua Shen, and G.~Zhang.
\newblock Show, attend and read: A simple and strong baseline for irregular
  text recognition.
\newblock In \emph{AAAI}, 2019.

\bibitem[Li et~al.(2020)Li, Yin, Li, Hu, Zhang, Zhang, Wang, Hu, Dong, Wei,
  Choi, and Gao]{li2020oscar}
Xiujun Li, Xi~Yin, Chunyuan Li, Xiaowei Hu, Pengchuan Zhang, Lei Zhang, Lijuan
  Wang, Houdong Hu, Li~Dong, Furu Wei, Yejin Choi, and Jianfeng Gao.
\newblock Oscar: Object-semantics aligned pre-training for vision-language
  tasks.
\newblock \emph{ECCV 2020}, 2020.

\bibitem[Lin et~al.(2014)Lin, Maire, Belongie, Bourdev, Girshick, Hays, Perona,
  Ramanan, Doll{\'{a}}r, and Zitnick]{ms_coco}
Tsung{-}Yi Lin, Michael Maire, Serge~J. Belongie, Lubomir~D. Bourdev, Ross~B.
  Girshick, James Hays, Pietro Perona, Deva Ramanan, Piotr Doll{\'{a}}r, and
  C.~Lawrence Zitnick.
\newblock Microsoft {COCO:} common objects in context.
\newblock \emph{CoRR}, abs/1405.0312, 2014.
\newblock URL \url{http://arxiv.org/abs/1405.0312}.

\bibitem[Litman et~al.(2020)Litman, Anschel, Tsiper, Litman, Mazor, and
  Manmatha]{scatter}
Ron Litman, Oron Anschel, Shahar Tsiper, Roee Litman, Shai Mazor, and
  R.~Manmatha.
\newblock Scatter: Selective context attentional scene text recognizer.
\newblock In \emph{2020 IEEE/CVF Conference on Computer Vision and Pattern
  Recognition (CVPR)}, pages 11959--11969, 2020.
\newblock \doi{10.1109/CVPR42600.2020.01198}.

\bibitem[Liu et~al.(2016)Liu, Chen, Wong, Su, and Han]{STAR-Net}
Wei Liu, Chaofeng Chen, Kwan-Yee~K. Wong, Zhizhong Su, and Junyu Han.
\newblock Star-net: A spatial attention residue network for scene text
  recognition.
\newblock In Edwin R.~Hancock Richard C.~Wilson and William A.~P. Smith,
  editors, \emph{Proceedings of the British Machine Vision Conference (BMVC)},
  pages 43.1--43.13. BMVA Press, September 2016.
\newblock ISBN 1-901725-59-6.
\newblock \doi{10.5244/C.30.43}.
\newblock URL \url{https://dx.doi.org/10.5244/C.30.43}.

\bibitem[Liu et~al.(2019)Liu, Meng, and Pan]{Liu_STR_survey}
Xiyan Liu, Gaofeng Meng, and Chunhong Pan.
\newblock Scene text detection and recognition with advances in deep learning:
  a survey.
\newblock \emph{International Journal on Document Analysis and Recognition
  (IJDAR)}, 22:\penalty0 143--162, 2019.

\bibitem[Long et~al.(2020)Long, He, and Yao]{Long_STR_survey}
Shangbang Long, Xin He, and C.~Yao.
\newblock Scene text detection and recognition: The deep learning era.
\newblock \emph{International Journal of Computer Vision}, 129:\penalty0
  161--184, 2020.

\bibitem[Loshchilov and Hutter(2017)]{AdamW}
Ilya Loshchilov and Frank Hutter.
\newblock Fixing weight decay regularization in adam.
\newblock \emph{CoRR}, abs/1711.05101, 2017.
\newblock URL \url{http://arxiv.org/abs/1711.05101}.

\bibitem[Lu et~al.(2021)Lu, Yu, Qi, Chen, Gong, Xiao, and Bai]{Lu2021MASTER}
Ning Lu, Wenwen Yu, Xianbiao Qi, Yihao Chen, Ping Gong, Rong Xiao, and Xiang
  Bai.
\newblock {MASTER}: Multi-aspect non-local network for scene text recognition.
\newblock \emph{Pattern Recognition}, 2021.

\bibitem[Mishra et~al.(2012)Mishra, Alahari, and Jawahar]{Mishra_BMVC}
Anand Mishra, Karteek Alahari, and Cv~Jawahar.
\newblock Scene text recognition using higher order language priors.
\newblock In \emph{Proceedings of the British Machine Vision Conference}, pages
  127.1--127.11. BMVA Press, 2012.
\newblock ISBN 1-901725-46-4.
\newblock \doi{http://dx.doi.org/10.5244/C.26.127}.

\bibitem[Mohajerin and Waslander(2017)]{rnn_init}
Nima Mohajerin and Steven~L. Waslander.
\newblock State initialization for recurrent neural network modeling of
  time-series data.
\newblock In \emph{2017 International Joint Conference on Neural Networks
  (IJCNN)}, pages 2330--2337, 2017.
\newblock \doi{10.1109/IJCNN.2017.7966138}.

\bibitem[Phan et~al.(2013)Phan, Shivakumara, Tian, and Tan]{phan_STR_2013}
Trung~Quy Phan, Palaiahnakote Shivakumara, Shangxuan Tian, and Chew~Lim Tan.
\newblock Recognizing text with perspective distortion in natural scenes.
\newblock In \emph{2013 IEEE International Conference on Computer Vision},
  pages 569--576, 2013.
\newblock \doi{10.1109/ICCV.2013.76}.

\bibitem[Sabu and Das(2018)]{Sabu_OCR_survey}
Abin~M Sabu and Anto~Sahaya Das.
\newblock A survey on various optical character recognition techniques.
\newblock In \emph{2018 Conference on Emerging Devices and Smart Systems
  (ICEDSS)}, pages 152--155, 2018.
\newblock \doi{10.1109/ICEDSS.2018.8544323}.

\bibitem[Saluja et~al.(2019)Saluja, Maheshwari, Ramakrishnan, Chaudhuri, and
  Carman]{OCR-on-the-go}
Rohit Saluja, Ayush Maheshwari, Ganesh Ramakrishnan, Parag Chaudhuri, and Mark
  Carman.
\newblock Ocr on-the-go: Robust end-to-end systems for reading license plates
  amp; street signs.
\newblock In \emph{2019 International Conference on Document Analysis and
  Recognition (ICDAR)}, pages 154--159, 2019.
\newblock \doi{10.1109/ICDAR.2019.00033}.

\bibitem[Shi et~al.(2016)Shi, Wang, Lyu, Yao, and Bai]{shi2017}
B.~Shi, X.~Wang, P.~Lyu, C.~Yao, and X.~Bai.
\newblock Robust scene text recognition with automatic rectification.
\newblock In \emph{2016 IEEE Conference on Computer Vision and Pattern
  Recognition (CVPR)}, pages 4168--4176, Los Alamitos, CA, USA, jun 2016. IEEE
  Computer Society.
\newblock \doi{10.1109/CVPR.2016.452}.
\newblock URL \url{https://doi.ieeecomputersociety.org/10.1109/CVPR.2016.452}.

\bibitem[Shi et~al.(2017)Shi, Bai, and Yao]{shi2016}
B.~Shi, X.~Bai, and C.~Yao.
\newblock An end-to-end trainable neural network for image-based sequence
  recognition and its application to scene text recognition.
\newblock \emph{IEEE Transactions on Pattern Analysis and Machine
  Intelligence}, 39\penalty0 (11):\penalty0 2298--2304, nov 2017.
\newblock ISSN 1939-3539.
\newblock \doi{10.1109/TPAMI.2016.2646371}.

\bibitem[Singh et~al.(2019)Singh, Natarjan, Shah, Jiang, Chen, Parikh, and
  Rohrbach]{singh2019towards}
Amanpreet Singh, Vivek Natarjan, Meet Shah, Yu~Jiang, Xinlei Chen, Devi Parikh,
  and Marcus Rohrbach.
\newblock Towards vqa models that can read.
\newblock In \emph{Proceedings of the IEEE Conference on Computer Vision and
  Pattern Recognition}, pages 8317--8326, 2019.

\bibitem[Singh et~al.(2021)Singh, Pang, Toh, Huang, Galuba, and
  Hassner]{singh2021textocr}
Amanpreet Singh, Guan Pang, Mandy Toh, Jing Huang, Wojciech Galuba, and Tal
  Hassner.
\newblock Textocr: Towards large-scale end-to-end reasoning for
  arbitrary-shaped scene text.
\newblock 2021.

\bibitem[Vaswani et~al.(2017)Vaswani, Shazeer, Parmar, Uszkoreit, Jones, Gomez,
  Kaiser, and Polosukhin]{transformer}
Ashish Vaswani, Noam Shazeer, Niki Parmar, Jakob Uszkoreit, Llion Jones,
  Aidan~N. Gomez, undefinedukasz Kaiser, and Illia Polosukhin.
\newblock Attention is all you need.
\newblock In \emph{Proceedings of the 31st International Conference on Neural
  Information Processing Systems}, NIPS'17, page 6000–6010, Red Hook, NY,
  USA, 2017. Curran Associates Inc.
\newblock ISBN 9781510860964.

\bibitem[Veit et~al.(2016)Veit, Matera, Neumann, Matas, and
  Belongie]{veit2016cocotext}
Andreas Veit, Tomas Matera, Lukas Neumann, Jiri Matas, and Serge Belongie.
\newblock Coco-text: Dataset and benchmark for text detection and recognition
  in natural images.
\newblock In \emph{arXiv preprint arXiv:1601.07140}, 2016.
\newblock URL
  \url{http://vision.cornell.edu/se3/wp-content/uploads/2016/01/1601.07140v1.pdf}.

\bibitem[Wang and Hu(2017)]{Wang2017_GRCNN}
Jianfeng Wang and Xiaolin Hu.
\newblock Gated recurrent convolution neural network for ocr.
\newblock In \emph{NIPS}, 2017.

\bibitem[Wang et~al.(2011)Wang, Babenko, and Belongie]{Wang_STR_2011}
Kai Wang, Boris Babenko, and Serge Belongie.
\newblock End-to-end scene text recognition.
\newblock In \emph{Proceedings of the 2011 International Conference on Computer
  Vision}, ICCV '11, page 1457–1464, USA, 2011. IEEE Computer Society.
\newblock ISBN 9781457711015.
\newblock \doi{10.1109/ICCV.2011.6126402}.
\newblock URL \url{https://doi.org/10.1109/ICCV.2011.6126402}.

\bibitem[Yang et~al.(2017)Yang, He, Zhou, Kifer, and Giles]{Yang_2017}
Xiao Yang, Dafang He, Zihan Zhou, Daniel Kifer, and C.~Lee Giles.
\newblock Learning to read irregular text with attention mechanisms.
\newblock In \emph{Proceedings of the Twenty-Sixth International Joint
  Conference on Artificial Intelligence, {IJCAI-17}}, pages 3280--3286, 2017.
\newblock \doi{10.24963/ijcai.2017/458}.
\newblock URL \url{https://doi.org/10.24963/ijcai.2017/458}.

\bibitem[Ye and Doermann(2015)]{Ye_STR_survey}
Qixiang Ye and D.~Doermann.
\newblock Text detection and recognition in imagery: A survey.
\newblock \emph{IEEE Transactions on Pattern Analysis and Machine
  Intelligence}, 37:\penalty0 1480--1500, 2015.

\bibitem[Zhang et~al.(2021)Zhang, Li, Hu, Yang, Zhang, Wang, Choi, and
  Gao]{zhang2021vinvl}
Pengchuan Zhang, Xiujun Li, Xiaowei Hu, Jianwei Yang, Lei Zhang, Lijuan Wang,
  Yejin Choi, and Jianfeng Gao.
\newblock Vinvl: Making visual representations matter in vision-language
  models.
\newblock \emph{CVPR 2021}, 2021.

\end{thebibliography}
\end{document}


\maketitle
